\documentclass[dvipsnames,format=sigconf,nonacm]{acmart}

\usepackage{multirow}
\usepackage{subcaption}

\DeclareMathOperator{\sign}{sign}

\newcommand{\nt}[1]{\langle \text{#1} \rangle} 
\newcommand{\term}[1]{\text{#1}}               

\title{NERO-Net: A Neuroevolutionary Approach for the Design of Adversarially Robust CNNs}

\author{Inês Valentim}
\orcid{0000-0001-6018-1788}
\affiliation{%
\institution{University of Coimbra, CISUC/LASI, DEI}
\city{Coimbra}
\country{Portugal}
}
\email{valentim@dei.uc.pt}
\author{Nuno Antunes}
\orcid{0000-0002-6044-4012}
\affiliation{%
\institution{University of Coimbra, CISUC/LASI, DEI}
\city{Coimbra}
\country{Portugal}
}
\email{nmsa@dei.uc.pt}
\author{Nuno Lourenço}
\orcid{0000-0002-2154-0642}
\affiliation{%
\institution{University of Coimbra, CISUC/LASI, DEI}
\city{Coimbra}
\country{Portugal}
}
\email{naml@dei.uc.pt}


\begin{document}


\begin{abstract}
    Neuroevolution automates the complex task of neural network design but often ignores the inherent adversarial fragility of evolved models which is a barrier to adoption in safety-critical scenarios. While robust training methods have received significant attention, the design of architectures exhibiting intrinsic robustness remains largely unexplored.
    In this paper, we propose NERO-Net, a neuroevolutionary approach to design convolutional neural networks better equipped to resist adversarial attacks. Our search strategy isolates architectural influence on robustness by avoiding adversarial training during the evolutionary loop. As such, our fitness function promotes candidates that, even trained with standard (non-robust) methods, achieve high post-attack accuracy without sacrificing the accuracy on clean samples.
    We assess NERO-Net on CIFAR-10 with a specific focus on $L_\infty$-robustness. In particular, the fittest individual emerged from evolutionary search with 33\% accuracy against FGSM, used as an efficient estimator for robustness during the search phase, while maintaining 87\% clean accuracy. Further standard training of this individual boosted these metrics to 47\% adversarial and 93\% clean accuracy, suggesting inherent architectural robustness. Adversarial training brings the overall accuracy of the model up to 40\% against AutoAttack.
\end{abstract}


\maketitle

\section{Introduction}
\label{sec:introduction}
The ubiquity of Artificial Neural Networks (ANNs) means that the vulnerabilities of these models have also become the center of attention of malicious actors.
In the image domain, a critical vulnerability lies in adversarial examples \cite{goodfellow_etal_2015,szegedy_etal_2014}: inputs containing small perturbations designed to deceive ANNs and induce misclassifications. Humans are often immune to these attacks, since the perturbed samples tend to be visually indistinguishable from their clean counterparts.

This vulnerability has triggered an arms race, with adversarial training~\cite{madry_etal_2018} (i.e., augmenting training data with adversarial examples) emerging as the standard defense. While alternatives like adversarial detection~\cite{metzen_etal_2017} exist, research remains skewed toward robust learning and regularization~\cite{mok_etal_2021}, largely overlooking the potential of intrinsically robust architectures.

While NeuroEvolution (NE) automates complex architecture design, it has historically prioritized predictive performance over robustness. The limited literature addressing this gap often makes compromising decisions to manage computational costs. Specifically, the prevalent use of cell-based search spaces with fixed backbones~\cite{mok_etal_2021,geraeinejad_etal_2021,dong_etal_2025} restricts the emergence of novel topologies, while the integration of adversarial training ~\cite{guo_etal_2020} during the search conceals the architecture's contribution to robustness.

To address these limitations, we propose a \textbf{N}euro\textbf{E}volutionary approach for the design of adversarially \textbf{RO}bust artificial neural \textbf{Net}works (NERO-Net).
We introduce novel mechanisms at the evaluation level, employing a fitness function that co-optimizes adversarial and clean performance. Candidate solutions undergo standard training, ensuring that the resulting robustness is intrinsic to the architecture rather than a by-product of adversarial training.

Focusing on image classification tasks and the design of robust Convolutional Neural Networks (CNNs), we show NERO-Net's efficacy on CIFAR-10~\cite{krizhevsky_2009}. Under restricted evolutionary training budgets, the fittest individual achieved 33\% adversarial accuracy (FGSM adversary in $L_\infty$) and 87\% clean accuracy. Extended standard training improved these metrics to 47\% and 93\%, respectively. When further refined via adversarial training, the model reached $\sim$40\% robust accuracy against AutoAttack. Remarkably, the discovered architecture exhibits multi-threat robustness, resisting $L_2$-bounded perturbations despite being optimized solely for an $L_\infty$ threat model.

In summary, our contributions are:
\begin{itemize}
    \item We propose \textbf{NERO-Net}, a neuroevolutionary framework that employs specialized fitness evaluation mechanisms to co-optimize accuracy and adversarial robustness;
    \item We introduce a flexible genotypic representation that expands the search space by encoding \textbf{tunable computational blocks} and supporting \textbf{a wider spectrum of topological connectivity patterns};
    \item We demonstrate that evolution can successfully navigate this complex landscape to discover architectures exhibiting \textbf{intrinsic robustness}, independent of specialized defenses.
\end{itemize}

The remainder of this paper is organized as follows. Section~\ref{sec:background-adversarial-robustness} introduces key concepts related to adversarial robustness and Section~\ref{sec:related work} reviews related work. Section~\ref{sec:nero-net} details the proposed approach. The experimental setup and the main findings of our experiments are presented and discussed in Sections~\ref{sec:evolutionary-search} and~\ref{sec:post-evaluation}, focusing on the evolutionary search and on further training after evolution, respectively. Section~\ref{sec:conclusion} concludes the paper and addresses future work.

\section{Adversarial Robustness}
\label{sec:background-adversarial-robustness}
In the image domain, an adversarial example $x^{adv}$~\cite{goodfellow_etal_2015,szegedy_etal_2014} is typically generated by adding small $L_{p}$-norm perturbations to a benign sample $x$, such that $\|x - x^{adv}\|_{p} \leq \epsilon$, where $\epsilon$ is the perturbation budget and usually $p \in \left\{0, 1, 2, \infty \right\}$~\cite{carlini_etal_2019,dong_etal_2018}. Although $x^{adv}$ is similar to the clean sample, the model gives it a highly different (and incorrect) prediction~\cite{szegedy_etal_2014,goodfellow_etal_2016}.

Considering an untargeted setting (a sample is misclassified as any incorrect class~\cite{carlini_wagner_2017a}) where the attacker has full access to the model, the Fast Gradient Sign Method (FGSM)~\cite{goodfellow_etal_2015} is a one-step gradient-based attack defined as follows for the $L_\infty$-norm:
$$
x^{adv} = x + \epsilon \cdot \sign \left(\nabla_{x} L(x, y)\right)
$$
where $y$ is the true label, $\nabla_{x} L(x, y)$ is the gradient of the cross-entropy (CE) loss with respect to the clean image and $x^{adv}$ is clipped to the valid data range.
The Projected Gradient Descent (PGD) method~\cite{madry_etal_2018} is an iterative, and stronger, variant of FGSM with smaller update steps. The starting point of the attack is usually a randomly perturbed sample (bounded by $\epsilon$) around the original input $x$~\cite{madry_etal_2018}. The APGD method~\cite{croce_hein_2020} further extends the PGD attack to progressively reduce the step size in an automated way, based on how the optimization is proceeding. A momentum term is also added, allowing the previous update to influence the current one.

These attacks can be adapted to the $L_{2}$-norm~\cite{metzen_etal_2017}: the clipping operation is changed to the projection onto the $L_{2}$-ball, and each update step follows the direction of the normalized gradient, i.e., $\frac{\nabla_{x} L(x, y)}{\left\|\nabla_{x} L(x, y)\right\|_{2}}$. A targeted version of the attacks can also be obtained by minimizing the loss for a specific target class, instead of maximizing it for the true class.

Heuristic approaches that approximate the robustness of a model by performing adversarial attacks are the standard evaluation practice~\cite{croce_etal_2021}, since exact computations are usually intractable~~\cite{carlini_etal_2019,croce_etal_2021}.
AutoAttack (AA)~\cite{croce_hein_2020} stands out among these approaches and tests adversarial robustness through an ensemble of diverse attacks: an untargeted APGD attack on the CE loss, an APGD attack on the targeted version of the Difference of Logits Ratio (DLR) loss, a targeted Fast Adaptive Boundary (FAB) attack~\cite{croce_hein_2020a}, as well as a Square attack~\cite{andriushchenko_etal_2020}. It is adopted by the RobustBench~\cite{croce_etal_2021} benchmark, which uses standardized evaluation methodologies to keep track of the progress made in adversarial robustness.

\section{Related Work}
\label{sec:related work}

There is a body of work in the literature that specifically explores the relationship between architecture and adversarial robustness \cite{huang_etal_2021,huang_etal_2023,peng_etal_2023}. That is the case of the work by~\citeauthor{huang_etal_2021}~\cite{huang_etal_2021}, which focuses on evaluating the impact of network width and depth on the robustness of a wide residual network.
Their results suggest that reduced width and depth at the last stage of the model can be beneficial. Another work \cite{huang_etal_2023} designs a family of adversarially robust residual networks (RobustResNets) after studying the contribution of several architectural components, both at the block level (e.g., layer parameterization and order) and the network scaling level (e.g., depth and width of each block in the network), to the robustness of the models.

However, a flaw of these studies is their focus on adversarially trained networks~\cite{madry_etal_2018}, making it difficult to assess the true role of architectural patterns and choices in the intrinsic robustness of models. Furthermore, they tend to analyze a single family of ANNs, with residual networks being the prevalent choice.

Considering an adversarial training setting, the role played by the choice of activation function \cite{xie_etal_2021,dai_etal_2022,peng_etal_2023}, specifically smooth and parameterized activation functions, and the inclusion of batch normalization layers \cite{wang_etal_2022} has also been studied. \citeauthor{huang_etal_2023}~\cite{huang_etal_2023} further argue that the benefits associated with a particular activation function depend on other design choices regarding the training setup (e.g., weight decay).

There is also a research line that applies NAS to automate the search for adversarially robust architectures~\cite{kotyan_vargas_2020a,geraeinejad_etal_2021,mok_etal_2021,guo_etal_2020,dong_etal_2025,sinn_etal_2019}. However, a significant portion relies on cell-based search spaces \cite{mok_etal_2021,guo_etal_2020,dong_etal_2025}. While efficient, these constrained spaces hinder the emergence of novel architectural patterns. Furthermore, these studies often rely on adversarial training \cite{guo_etal_2020,sinn_etal_2019}, once again making it difficult to isolate the architecture's contribution to robustness.

Among NE-based approaches, R-NAS~\cite{kotyan_vargas_2020a} and RoCo-NAS~\cite{geraeinejad_etal_2021} explicitly optimize robustness alongside clean accuracy (and computational complexity in RoCo-NAS). Both estimate adversarial robustness using pre-generated adversarial examples (transfer-based attacks) to minimize computational overhead. We argue that this surrogate metric is insufficient compared to direct evaluation strategies. Additionally, the final validation protocol in R-NAS remains ambiguous.

\citeauthor{devaguptapu_etal_2021}~\cite{devaguptapu_etal_2021} and \citeauthor{valentim_etal_2022}~\cite{valentim_etal_2022} share the same goal of evaluating and comparing the adversarial robustness of models designed both by human experts and via NAS approaches, with an emphasis on NE in the case of \citeauthor{valentim_etal_2022}~\cite{valentim_etal_2022}. In both works, the models under evaluation are not explicitly designed to resist adversarial attacks. While \citeauthor{devaguptapu_etal_2021}~\cite{devaguptapu_etal_2021} only consider $L_{\infty}$-robustness, \citeauthor{valentim_etal_2022}~\cite{valentim_etal_2022} also include $L_2$-robustness in their experimental campaign.

\citeauthor{jung_etal_2023}~\cite{jung_etal_2023} come closer to providing a standardized way to assess the robustness of candidate solutions in a NAS-based setting by extending the NAS-Bench-201 benchmark~\cite{dong_yang_2019} with several metrics, including some related to $L_\infty$-robustness. The authors use the pre-trained models from the baseline benchmark, which were optimized with standard (non-robust) training methods. Although there are different use cases for this extended version of the benchmark, they are limited to approaches that adopt the same cell-based search space and pre-defined operation set.

\section{NERO-Net}
\label{sec:nero-net}
In this work, we propose NERO-Net (\textbf{N}euro\textbf{E}volution of adversarially \textbf{RO}bust artificial neural \textbf{Net}works), a framework for designing intrinsically robust CNNs. Our method builds upon Fast-DENSER~\cite{assuncao_etal_2021}, an extension of the original DENSER algorithm~\cite{assuncao_etal_2019}. The following sections detail the modifications made to tailor Fast-DENSER for this task, focusing primarily on the redesign of the fitness function and the evaluation strategy for candidate solutions.

\subsection{Overview of DENSER}
\label{ssec:denser}
DENSER relies on a hierarchical two-level representation for each individual. The outer level defines the ANN's macro-structure as an ordered sequence of evolutionary units (e.g., layers). The inner level governs the parameterization of these units using a schema based on Dynamic Structured Grammatical Evolution (DSGE), encoding parameters as a sequence of grammatical derivation steps. Unlike traditional DSGE implementations, the inner level of DENSER encodes real-valued parameters directly, allowing continuous optimization within the grammatical framework.

At the outer level, units represent ANN layers or auxiliary components, like learning policies. Each unit is categorized by a type, where contiguous units of the same type form a module. These types correspond to non-terminal symbols in the inner level context-free grammar, ensuring a consistent structural definition across the two-level representation.

Overall, the search space of a DENSER-based approach and the valid genotype structure is dictated by the configuration of the outer level and the grammatical rules of the inner level. The outer level structure is specified through a sequence of 3-element tuples, each representing a distinct module. These tuples enforce structural constraints by defining the type of the module's units (linked to a grammar non-terminal) and its size bounds (minimum and maximum number of units). Finally, the inner level resolves the specific parameters via the grammar's production rules.

Fast-DENSER instantiates the DENSER framework using a $(1 + \lambda)$ Evolutionary Strategy (ES)~\cite{assuncao_etal_2021}. It further refines the architecture by incorporating a distinct genotypic level for layer connectivity, where units explicitly reference their inputs. The associated variation operators encompass structural changes (adding, replicating, or removing units, and adding or removing connections), parameter tuning (grammatical and real-value mutations), and training time adjustments.

\subsection{Fitness Function}
\label{ssec:fitness}
We aim at designing CNNs that balance high clean accuracy with intrinsic adversarial robustness. Similarly to DENSER, individuals are optimized via standard training procedures. However, contrary to the original DENSER formulation, which relies solely on standard performance metrics, our quality assessment uses $F_\beta$, a fitness function that explicitly combines clean accuracy ($C$) with adversarial accuracy ($A$). Formally, this resembles the weighted Harmonic Robustness Score (HRS)~\cite{devaguptapu_etal_2021}, where the parameter $\beta$ controls the weight given to preserving performance on benign data over perturbed data. As such, it is defined as:
\begin{equation*}
    \setlength{\abovedisplayskip}{2pt}
    \setlength{\belowdisplayskip}{2pt}
    F_\beta = (1 + \beta^2) \cdot \frac{C \times A}{C + \beta^2 \cdot A}
\end{equation*}
\begin{align*}
    \setlength{\abovedisplayskip}{2pt}
    \setlength{\belowdisplayskip}{2pt}
    C = \frac{\sum_{i=1}^{N} I(y_i, y_i^{clean})}{N} = \frac{N_c}{N} && A = \frac{\sum_{i=1}^{N_c} I(y_i, y_i^{adv})}{N_c}
\end{align*}
where $N$ is the number of clean samples, $N_c$ is the number of clean samples that are correctly classified, $y^{clean}$ and $y^{adv}$ are the predicted labels for a clean or an adversarial sample, respectively, and $I(y,\hat{y})$ is an indicator function that is 1 when $y = \hat{y}$.

The fitness evaluation is inherently computationally intensive due to the requirement of training individuals via backpropagation. Consequently, we employ the FGSM attack to estimate adversarial accuracy efficiently. We acknowledge that FGSM is a single-step attack and may yield an optimistic estimation of robustness. However, since candidate solutions do not undergo explicit adversarial training, stronger attacks would likely drive adversarial accuracy to near-zero values across the entire population. Such a performance degradation would eliminate selective pressure, rendering the evolutionary algorithm unable to distinguish between individuals with varying degrees of intrinsic robustness.

Bearing in mind that only the correctly classified clean samples are taken into consideration when generating the adversarial examples for an individual, the number of attacked samples typically varies among the individuals in the population.

\paragraph{Warm-up Period}
To facilitate the emergence of viable architectures, our approach incorporates a \textit{warm-up phase}. During this initial period, adversarial examples are not considered, and the fitness function is reduced solely to clean accuracy ($C$). This strategy allows the evolutionary search to establish a baseline of predictive performance before introducing the constraint of adversarial robustness. The duration of this phase is adaptive, governed by a hyper-parameter $\tau$. Specifically, at the end of each generation, we evaluate the population's mean fitness, and once it meets or exceeds the threshold $\tau$, the optimization objective permanently transitions to the composite function $F_\beta$.

\paragraph{Detection of Ill-Fitted Individuals}
Preliminary experiments revealed that the evolutionary search occasionally converged toward degenerate solutions, which we term \textit{ill-fitted individuals}. These manifested in two primary forms: (1) numerical instability, characterized by exploding loss values during training; and (2) trivial classification, where models predicted a single class for the vast majority of samples. The latter created a ``robustness trap'': since adversarial accuracy is conditional on correct initial classification, a model that blindly guesses one class may achieve artificially high robustness on that specific subset.

To mitigate this, we implemented a validity constraint. An individual is flagged as ill-fitted if:
\begin{enumerate}
    \item Its training history contains non-finite loss values (NaN or infinity); or
    \item The prediction frequency for any single class exceeds a threshold of $100 \times \left( 1 - \frac{1}{N_{classes}} \right)$\%.
\end{enumerate}
If either condition is met, the expensive adversarial evaluation is bypassed, and the quality of the individual penalized.

\subsection{Representation and Variation Operators}
\label{ssec:representation-and-operators}

\paragraph{Blocks of Layers}
In DENSER, a one-to-one mapping between non-terminal symbols and ANN layers originates a highly flexible, but excessively large search space. To address this, we leverage the observation that state-of-the-art CNNs recurrently rely on composite blocks of operations --- such as Batch Normalization (BN), rectified linear unit (ReLU) activation, and convolution (conv) --- rather than isolated layers~\cite{huang_etal_2017}. The operation ordering varies across well-established architectures~\cite{he_etal_2016,he_etal_2016a}.

Consequently, we introduce new non-terminal symbols (and new layer types --- \texttt{convblock} and \texttt{poolblock}) that encode these layer groupings directly into the grammar (see Figure~\ref{fig:grammar-snippet}). The \texttt{conv-block} symbol encapsulates a convolutional layer, an activation function, and an optional BN layer, preserving all the standard convolution parameters (e.g., kernel size, filter count) and letting the grammatical rules determine the internal ordering of the operations.

\begin{figure}
    \setlength{\abovecaptionskip}{5pt}
    \footnotesize
    \centering
    \begin{minipage}{\columnwidth}
        \begin{alignat*}{2}
        \nt{conv-block}   & ::= \; && \term{layer:convblock} \; \nt{act-position} \; \nt{activation} \; \nt{bn-position} \\
                          &        && \term{[num-filters,int,1,32,256]} \; \term{[filter-shape,int,1,1,5]} \\
                          &        && \term{[stride,int,1,1,3]} \; \nt{padding} \; \nt{bias} \\
        \nt{act-position} & ::= \; && \term{act-pos:preconv} \mid \term{act-pos:postconv} \\
        \nt{bn-position}  & ::= \; && \term{bn:pre} \mid \term{bn:mid} \mid \term{bn:post} \mid \term{bn:none}
        \end{alignat*}
    \end{minipage}
    \caption{Snippet of a grammar with new rules for the \texttt{convblock} layer type.}
    \label{fig:grammar-snippet}
\end{figure}

No changes to the genotype-to-phenotype mapping are required; only the model compilation is updated. Internally, the block's layers are connected sequentially. Then, the first and last layers serve as the interface to the global network topology, receiving and aggregating the block's inputs, and acting as the block's output, respectively.

\paragraph{Connections between Layers}
In Fast-DENSER, each layer can aggregate inputs from multiple preceding layers. This connectivity is governed by a parameter called ``levels back'', which establishes a maximum range for valid connections. Indirectly, it also controls the maximum number of inputs that each layer can have. Furthermore, by being module-specific, this parameter allows different architectural blocks to enforce distinct local connectivity constraints.

\begin{figure}
    \setlength{\abovecaptionskip}{5pt}
    \setlength{\belowcaptionskip}{-8pt}
    \centering
    \includegraphics[width=\columnwidth]{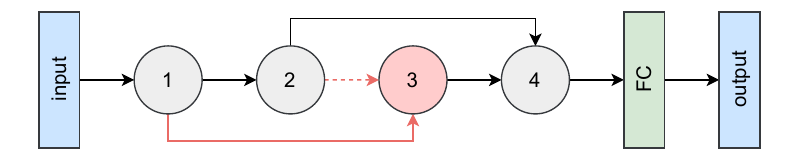}
    \caption{An illustration of the connections between layers, assuming the number of levels back is 2. The original implementation does not allow layer 3 to have layer 1 as its only input (red connection). Layer 2 would always have to be an input of layer 3 (dashed red connection).}
    \label{fig:real-skip-connections}
\end{figure}

However, in the original implementation, each layer must always be connected to the previous one. Consequently, it would not be possible to design a CNN like the one shown in Figure~\ref{fig:real-skip-connections}, since layer 3 would also have to be connected to layer 2 (dashed red connection).
We no longer impose this constraint, this way allowing for \textit{skip connections} to occur (red connection between layer 1 and layer 3).

To support these new connectivity patterns, we updated the initialization procedure and the mutation operators. Every new unit must have at least one input connection from its valid predecessors, without being restricted to the immediate previous layer.
Consequently, mutations for adding or removing connections treat the link to the immediate predecessor under the same probabilistic constraints as any other viable candidate.
To prevent ``dead ends'' in the architecture, we implemented a validity check: if any intermediate layer lacks outbound connections, we force a connection to a randomly selected valid successor (respecting the look-back limit). This repair mechanism is triggered during initialization and post-mutation. Finally, the outer level definition of modules is extended to 4-element tuples, where the additional parameter specifies whether the module permits the new skip connections or is restricted to the legacy connectivity scheme.

\paragraph{New Types for Real-Valued Parameters}
At the inner level representation, real-valued parameters appear as 5-element tuples in the grammar, containing the parameter's name and type (originally, either integer or float), the number of values to generate, the minimum and the maximum possible value that can be generated. We added \texttt{int\_power2} and \texttt{int\_power10} as new types: for both cases, we randomly generate as many integers as defined in the tuple, within the valid range of values. The parameter is then mapped to either 2 or 10 to the power of the generated integer(s). For decimal numbers, the search space can be greatly reduced by using these new parameter types.

\paragraph{Training Time Mutation}
In the latest DENSER iteration, the training duration is dynamic and can be extended through two distinct mechanisms. First, a mutation operator can incrementally increase an individual's training budget by a fixed default interval. Second, the selection operator can autonomously extend the training time to ensure a fair comparison between high-performing candidates that have been trained for different durations.

While beneficial for accuracy, this unbounded growth can lead to excessive computational costs. To mitigate this, we introduce an upper bound on training time. If the mutation to extend training is triggered but the individual has already reached this cap, the operation is effectively ``short-circuited'': the mutation is ignored, and, in a strict resetting protocol, the training time is reverted to its default value and other genetic modifications can take place.

\paragraph{Initial Population}
\label{ssec:initial-population}
The initial population of Fast-DENSER is randomly generated: $\lambda$ individuals are created with up to a maximum pre-defined number of layers that are randomly connected, according to the configured set of hyper-parameters. Additionally, any other outer level units (e.g., unit encoding the learning strategy) are randomly initialized. Following this approach, the first generation of individuals most likely exhibits poor fitness.

We add the possibility of using a known architecture as a seed for the evolutionary process, but the corresponding genotype must be defined manually. The goal is to speed up evolution by introducing genetic material associated with a promising candidate solution early on in the search process.
When choosing this option, the initial population consists of an individual with the given architecture, plus $\lambda - 1$ individuals created from it through mutation. We maintain the random initialization of other outer level units.

\section{Evolutionary Search}
\label{sec:evolutionary-search}
We evaluate NERO-Net on the design of CNNs for CIFAR-10 \cite{krizhevsky_2009}. We focus on $L_\infty$-robustness for several reasons: a brief perusal of the RobustBench leaderboards\footnote{\url{https://robustbench.github.io}} reveals that this threat model receives the most attention, with almost 100 entries; and this also seems to be the preferred threat model for works that combine NAS and adversarial robustness \cite{mok_etal_2021,dong_etal_2025,guo_etal_2020}. In sum, this choice makes it easier for our work to be compared with the existing literature.

The specification of the outer-level structure and the design of the grammar are partly motivated by the results of prior works that evaluate the adversarial robustness of models designed via NE approaches~\cite{devaguptapu_etal_2021,valentim_etal_2022}. According to these studies, the NSGA-Net~\cite{lu_etal_2019} model from a macro search space (the connections between blocks of layers are the sole target of evolution) seems to show some degree of resistance to simpler attacks like FGSM.
Therefore, our choices aimed at designing a search space that would encompass the architecture of the NSGA-Net model.
We can view our search space as a less restricted version of the NSGA-Net's macro search space, since we do not define a fixed macro-structure of the architecture and some layer parameters still undergo evolution, instead of also being defined apriori (e.g., number of convolution filters). However, we have to deal with a larger search space and a higher number of candidate solutions.

We use the grammar presented in Figure~\ref{fig:grammar-neronet} in conjunction with the following outer level structure:

\begin{center}
    [(\texttt{stem}, 0, 1, false), (\texttt{features}, 1, 30, true),\\
    (\texttt{last-transition}, 0, 1, false), (\texttt{classification}, 0, 5, false),\\
    (\texttt{softmax}, 1, 1, false), (\texttt{learning}, 1, 1)]
\end{center}

\begin{figure}
    \setlength{\abovecaptionskip}{5pt}
    \footnotesize
    \centering
    \begin{minipage}{\columnwidth}
\begin{alignat*}{2}
    \nt{stem}                & ::= \; && \term{layer:convblock} \; \term{act-pos:postconv} \; \term{act:linear} \\
                             &        && \term{bn:none} \; \term{[num-filters,int,1,16,256]}\\
                             &        && \term{filter-shape:3} \; \term{stride:1} \; \nt{padding} \; \term{bias:False} \\
    \nt{features}            & ::= \; && \nt{macro-node} \mid \nt{macro-node} \mid \nt{transition-block} \\
    \nt{macro-node}          & ::= \; && \term{layer:macro-node} \; \nt{node-activation} \\
                             &        && \term{[num-filters,int,1,16,256]} \; \term{filters-mult:4} \\
    \nt{node-activation}     & ::= \; && \term{act:relu} \mid \term{act:swish} \\
    \nt{transition-block}    & ::= \; && \term{layer:transition} \; \term{act-pos:postconv} \; \nt{node-activation} \\
                             &        && \term{conv-bn:mid} \; \term{[num-filters,int,1,16,256]} \; \term{conv-filter-shape:1} \\
                             &        && \term{conv-stride:1} \; \term{conv-padding:valid} \; \term{conv-bias:False} \\
                             &        && \nt{pooling-type} \; \term{pool-kernel-size:2} \\
                             &        && \term{pool-stride:2} \; \term{pool-padding:valid} \; \term{pool-bn:none} \\
    \nt{last-transition}     & ::= \; && \term{layer:transition} \; \term{act-pos:postconv} \; \nt{node-activation} \\
                             &        && \term{conv-bn:mid} \; \term{[num-filters,int,1,16,256]} \; \term{conv-filter-shape:1} \\
                             &        && \term{conv-stride:1} \; \term{conv-padding:valid} \; \term{conv-bias:False} \\
                             &        && \nt{pooling-type} \; \term{[pool-kernel-size,int,1,2,7]} \\
                             &        && \term{pool-stride:1} \; \term{pool-padding:valid} \; \term{pool-bn:none} \\
    \nt{classification}      & ::= \; && \term{layer:fc} \; \nt{act-function} \; \term{[num-units,int,1,128,2048]} \; \nt{bias} \\
    \nt{pooling-type}        & ::= \; && \term{pooling:avg} \mid \term{pooling:max} \\
    \nt{padding}             & ::= \; && \term{padding:same} \mid \term{padding:valid} \\
    \nt{bias}                & ::= \; && \term{bias:True} \mid \term{bias:False} \\
    \nt{act-function}        & ::= \; && \term{act:relu} \mid \term{act:relu} \mid \term{act:swish} \mid \term{act:swish} \mid \term{act:sigmoid} \\
    \nt{softmax}             & ::= \; && \term{layer:fc} \; \term{act:softmax} \; \term{num-units:10} \; \term{bias:True} \\
    \nt{learning}            & ::= \; && \nt{optimizer-algo} \; \term{[early\_stop,int,1,5,20]} \\
                             &        && \term{[batch\_size,int\_power2,1,5,9]} \; \term{epochs:10000} \\
    \nt{optimizer-algo}      & ::= \; && \nt{gradient-descent} \mid \nt{rmsprop} \mid \nt{adam} \\
    \nt{gradient-descent}    & ::= \; && \term{learning:gradient-descent} \; \nt{learning-rate} \\
                             &        && \term{[momentum,float,1,0.68,0.99]} \; \nt{nesterov} \\
    \nt{nesterov}            & ::= \; && \term{nesterov:True} \mid \term{nesterov:False} \\
    \nt{adam}                & ::= \; && \term{learning:adam} \; \nt{learning-rate} \\
                             &        && \term{[beta1,float,1,0.5,1]} \; \term{[beta2,float,1,0.5,1]} \\
    \nt{rmsprop}             & ::= \; && \term{learning:rmsprop} \; \nt{learning-rate} \; \term{[rho,float,1,0.5,1]} \\
    \nt{learning-rate}       & ::= \; && \term{[lr,int\_power10,1,-6,-1]} \; \term{[decay,int\_power10,1,-6,-3]}
\end{alignat*}
    \end{minipage}
    \caption{Grammar used in NERO-Net.}
    \label{fig:grammar-neronet}
    \vspace{-8pt}
\end{figure}

Our grammar encodes CNNs with a DenseNet-like backbone~\cite{huang_etal_2017}, with an optional $3 \times 3$ convolutional stem that increases the number of initial channels.
The core feature extraction portion of the network relies on two unit types, where \texttt{macro-nodes} are selected with higher probability than \texttt{transition-blocks} to mimic the structural phases of NSGA-Net.
The primary computational units, \texttt{macro-nodes}, consist of a bottleneck block ($1 \times 1$ convolution with $4 \times$ expansion) followed by a $3 \times 3$ convolutional block. Both blocks adopt a fixed BN-activation-conv order, with evolvable parameters for filter count and activation function (ReLU or Swish/SiLU). These are complemented by \texttt{transition-blocks}, which reduce spatial resolution via a $1 \times 1$ convolutional block (conv-BN-activation) followed by stride-2 pooling; here, the pooling type (average/max), filter count, and activation are free parameters.
The \texttt{features} units are the only to support the new skip connections with a look-back depth of 5 layers.
Following feature extraction, an optional \texttt{last-transition} unit performs global pooling (stride 1) with an evolvable kernel size. Unlike standard single-layer heads, the classification head can have up to 5 fully-connected units (fully-connected layer + activation) before the final \texttt{softmax} layer, a choice motivated by previous DENSER results~\cite{assuncao_etal_2019}. Finally, the \texttt{learning} unit optimizes training hyper-parameters, including batch size, early stopping, and the optimizer (SGD, Adam, or RMSProp) with an evolvable inverse time decay learning rate schedule.

\subsection{Experimental Setup}
\label{ssec:experimental-setup}
In what follows, we detail the experimental setup used for running NERO-Net, with the main parameters summarized in Table~\ref{tab:experimental-setup}. The new threshold $\tau$ and weight factor $\beta$ were set after analyzing the results of preliminary experiments, while the remaining parameters were set according to general configurations from DENSER \cite{assuncao_etal_2019} and Fast-DENSER \cite{assuncao_etal_2019a}. Nevertheless, we use a higher mutation rate for replicating a layer as an attempt to stir the search towards modular architectures, particularly regarding \texttt{macro-node} blocks. We also lower the  probability of applying the training time mutation as an effort to enforce a better exploration of the search space, since there is a higher chance of producing offspring that are not exact copies of the parent's architecture. Additionally, we use the architecture of the NSGA-Net model from their macro search space as a seed for the initial population, as described in Section~\ref{ssec:initial-population}.

\begin{table}
    \caption{NERO-Net experimental parameters.}
    \label{tab:experimental-setup}
    \footnotesize
    \begin{tabular}{lc}
    \toprule
    \textbf{Parameter}                           & \textbf{Value}                 \\
    \midrule
    Number of runs                               & 5                              \\
    Number of generations                        & 100                            \\
    Number of offspring ($\lambda$)              & 4                              \\
    Add layer mutation rate                      & 25\%                           \\
    Replicate layer mutation rate                & 35\%                           \\
    Remove layer mutation rate                   & 25\%                           \\
    Add connection mutation rate                 & 15\%                           \\
    Remove connection mutation rate              & 15\%                           \\
    Grammatical mutation rate (layer unit)       & 15\%                           \\
    Grammatical mutation rate (learning unit)    & 30\%                           \\
    Gaussian mutation (float)                    & $\mu = 0.00$, $\sigma = 0.15 $   \\
    Increase training time mutation rate         & 10\%                           \\
                                                 &                                \\
    Evolutionary training set                    & 43000 instances                \\
    Evolutionary control set                     & 3500 instances                 \\
    Fitness evaluation set                       & 3500 instances                 \\
                                                 &                                \\
    Default training time                        & 10 minutes                     \\
    Maximum training time                        & 30 minutes                     \\
                                                 &                                \\
    Warm-up period threshold ($\tau$)            & 0.80                           \\
    Fitness function weight factor ($\beta$)     & 4                              \\
    \bottomrule
    \end{tabular}
    \vspace{-10pt}
\end{table}

During each evolutionary run, we evaluate 400 individuals. These do not necessarily encode 400 different architectures, since a child can have the same architecture of the parent, for instance, if the training time mutation is applied.

\paragraph{Dataset}
We use CIFAR-10 ($32 \times 32$ RGB images, 10 classes), normalizing pixel values to $[0,1]$. Following the partitioning scheme of \citeauthor{valentim_etal_2024}~\cite{valentim_etal_2024}, the standard test set (10,000 images) is strictly isolated from the search process. In each evolutionary run, the 50,000 training images are randomly divided into three subsets, 
using distinct seeds to ensure experimental independence.

\paragraph{Training Strategy}
Candidates are trained on the evolutionary training set using standard data augmentation (4-pixel padding, random horizontal flips, and cropping)~\cite{suganuma_etal_2017, assuncao_etal_2019}, while the control set regulates early stopping. A fixed L2 regularization ($5 \times 10^{-4}$) is applied exclusively to convolutional and fully-connected kernels.

\paragraph{Fitness Evaluation}
During evolution, candidates are evaluated on the fitness subset. Prior to reaching the warm-up threshold $\tau$, fitness is defined solely by clean accuracy. Once $\tau$ has been surpassed, we transition to the $F_\beta$ metric (with $\beta=4$, see Table~\ref{tab:experimental-setup}). Adversarial accuracy is computed using an FGSM adversary with perturbations bounded by $\epsilon = 8/255$ ($L_\infty$-norm) and a batch size of 128. This standard perturbation budget~\cite{croce_etal_2021} balances attack strength against the intrinsic vulnerability of undefended models, ensuring meaningful differentiation between candidates.

\subsection{Results and Analysis}
\label{ssec:evolution-results}
\begin{figure}[t!]
    \setlength{\abovecaptionskip}{5pt}
    \setlength{\belowcaptionskip}{-15pt}
    \centering
    \begin{subfigure}[b]{\columnwidth}
        \setlength{\abovecaptionskip}{1pt}
        \setlength{\belowcaptionskip}{1pt}
        \centering
        \includegraphics[width=0.85\columnwidth]{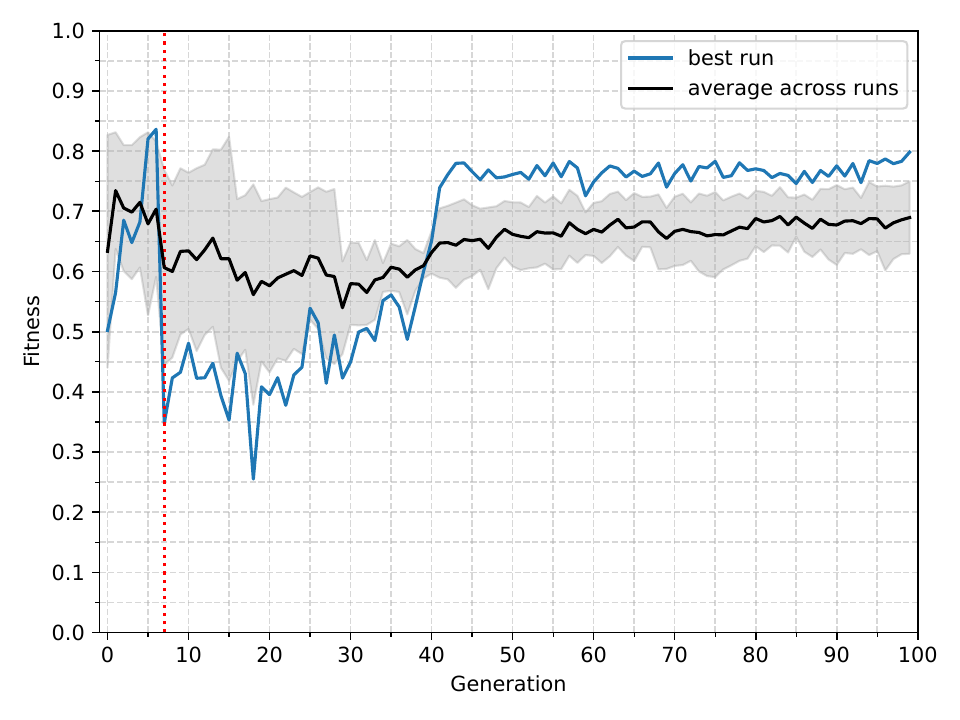}
        \caption{Fitness}
        \label{fig:best-fitness}
    \end{subfigure}
    \begin{subfigure}[b]{\columnwidth}
        \setlength{\abovecaptionskip}{1pt}
        \setlength{\belowcaptionskip}{1pt}
        \centering
        \includegraphics[width=0.85\columnwidth]{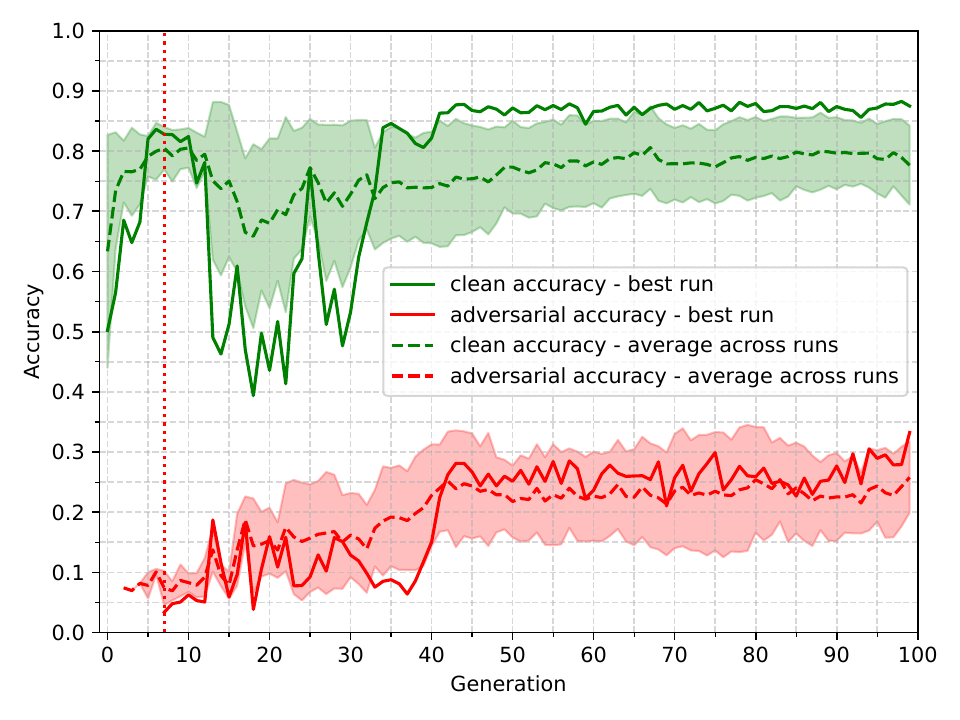}
        \caption{Clean Accuracy vs. Adversarial Accuracy}
        \label{fig:best-clean-adversarial-acc}
    \end{subfigure}
    \caption{Evolution of the best individual per generation. The red dotted lines mark the transition based on the $\tau$ threshold.}
    \label{fig:evolution-metrics}
\end{figure}

In this section, we present the results of the independent runs of NERO-Net. Figure~\ref{fig:best-fitness} shows the progress of the fitness of the best individual per generation, both averaged across all runs (gray area represents standard deviation) and for the run with the overall best individual. The red dotted line represents the first generation where the fitness function is $F_\beta$ (i.e., the threshold $\tau$ was met in the previous generation) during the best run.

The warm-up periods tend to be relatively short, which we attribute to the evolution not starting from randomly generated individuals, but rather from an architecture known to perform well on the CIFAR-10 classification task. However, we also observe a high variance up to, approximately, the 30th generation. In fact, for the run with the longest warm-up period, $\tau$ is met at generation 29.

The fitness of the best individual usually drops significantly right after the transition to the $F_\beta$ function. Keeping in mind that the duration of the warm-up period varies between runs, we can easily conclude that the high variance at the start of the evolutionary process is related to this phenomenon.

Figure~\ref{fig:best-fitness} also shows that the average fitness starts improving from around the 30th generation onward, but it does not seem to follow a monotonically increasing trend. Figure~\ref{fig:best-clean-adversarial-acc} shows how the clean and adversarial accuracy of the best individuals from each generation evolve throughout the search process.

We can see that the drop in fitness after $\tau$ is reached is mainly due to the low adversarial accuracy of the best individuals of the first post-transition generations. After a certain point in the evolutionary process, the clean accuracy of the best individuals stabilizes, with fitness fluctuations being mainly due to changes in adversarial accuracy.
In our experiments, we noticed that small genotypic changes lead to more unpredictable outcomes in terms of adversarial accuracy than in terms of clean accuracy. Thus, we hypothesize that incorporating adversarial accuracy in the objective function leads to more intricate fitness landscapes.

The clean accuracy of the best individuals of the last generations is, on average, around 80\%, which equals the value of $\tau$. This means that, at the end of the evolutionary process, the clean accuracy of these individuals is close to the mean population fitness before adversarial robustness is incorporated as an optimization goal.
Moreover, comparing the last generations with the early ones, the results suggest gains of more than 18 percentage points in adversarial accuracy.
Thus, it is safe to say that NERO-Net succeeds in finding CNNs that are inherently more robust, while remaining competitive in terms of predictive accuracy on clean data samples.

The best individual across all generations and evolutionary runs attains a fitness of 0.7983, a result of a clean accuracy of 87.49\% and an adversarial accuracy of 33.25\%. We reiterate that these measurements are obtained on data from the fitness evaluation set, without the inclusion of any adversarial defense.

This individual contains a convolutional stem, but not a final global pooling unit. The feature extraction portion of the network has 13 \texttt{macro-nodes} and 7 \texttt{transition-blocks} (5 average vs. 2 max pooling). The ratio between ReLU and SiLU in these blocks is almost 1:1. These evolutionary units take advantage of real skip connections, and unlike what happens with traditional models, some \texttt{macro-nodes} connect to others that precede the closest \texttt{transition-block}. The classification head before the final softmax layer has 4 fully-connected units (3 SiLU vs. 1 ReLU activation).

\section{Further Training and Evaluation}
\label{sec:post-evaluation}

While Fast-DENSER typically yields fully converged models, NERO-Net enforces a strict 30-minute training cap to ensure computational tractability (Section~\ref{ssec:representation-and-operators}). An analysis of the fittest individuals reveals that a majority (322/500) saturate this budget, averaging only $\approx$ 50 epochs. We hypothesize that the best evolved architecture requires extended training to reach its potential. Therefore, we evaluate the fittest individual using both prolonged standard training and complementary adversarial training.

\subsection{Experimental Setup}
As far as NERO-Net is concerned, we select the fittest individual (as given by the fitness on the fitness evaluation data subset) across all generations and evolutionary runs. Additionally, we also train and evaluate the NSGA-Net \cite{lu_etal_2019} model from the macro search space, whose architecture (i.e., the pre-defined macro-structure\footnote{\url{https://github.com/iwhalen/nsga-net/blob/master/models/macro\_models.py}} and the genotype of each phase\footnote{\url{https://github.com/iwhalen/nsga-net/blob/master/models/macro\_genotypes.py}}) is publicly available on the source code repository of NSGA-Net. Due to the stochastic nature of the process, we train each model 10 times, using different seeds for the random generators. We then run each adversarial attack 10 times, such that each trained model is used once per adversarial attack.

\paragraph{Dataset}
Following \citeauthor{valentim_etal_2024}~\cite{valentim_etal_2024}, we merge the original training set back together for these additional training runs, and evaluate the trained models on the original test set. As before, the only pre-processing applied to the data is to normalize the pixel values to the interval [0, 1].

\paragraph{Standard Training Setting}
For the standard (non-robust) training method, we use a stochastic gradient descent (SGD) optimizer with momentum (set to 0.90) for 350 epochs, with a batch size of 128. The initial learning rate is 0.025 and it is annealed to zero through a cosine decay schedule. We use a weight decay of $3 \times 10^{-4}$ and gradient norm clipping (threshold of 5.0). We apply the same data augmentation scheme used during the evolutionary search runs (padding, horizontal flipping, and random cropping).
This setting closely resembles how the NSGA-Net model is trained in the original work, hence its adoption here. One of the main differences resides in the fact that we do not include cutout \cite{devries_taylor_2017} in our data augmentation scheme.

\paragraph{Adversarial Training Setting}
For adversarial training, we consider a 7-step PGD adversary, with $\epsilon = 8/255$ and step size $\epsilon / 4$. We use an SGD optimizer with momentum (set to 0.90) for 200 epochs, with a batch size of 64. The initial learning rate is 0.10, and it is decayed by a factor of 0.1 at epochs 100 and 150. We use a weight decay of $1 \times 10^{-4}$ and gradient norm clipping (threshold of 5.0). As in the standard training setting, we apply the same data augmentation scheme used during the evolutionary runs.
This setting is commonly adopted by approaches that combine NAS and adversarial robustness \cite{feng_etal_2025,mok_etal_2021,guo_etal_2020}, which in turn tend to follow the adversarial training method proposed by \citeauthor{madry_etal_2018}~\cite{madry_etal_2018}.

\paragraph{Threat Models and Attacks}
We consider the threat model used during evolutionary search ($L_\infty$-norm with $\epsilon = 8/255$).
Additionally, we evaluate the models under a threat model with $L_2$-bounded perturbations ($\epsilon = 0.5$), to examine how they behave when presented with data perturbations not seen during evolution.
We run all the attacks with a batch size of 32. We use the TensorFlow implementations by the Adversarial Robustness Toolbox library~\cite{nicolae_etal_2019} for all attacks except AA, for which we use the original PyTorch implementation\footnote{\url{https://github.com/fra31/auto-attack}}.

\subsection{Results and Analysis}
We first evaluate the models trained with a standard (non-robust) method. Table~\ref{tab:results} shows the results for FGSM, PGD with 20 iterations (step size $\epsilon / 4$), and the standard version of AA.



We confirm our hypothesis that our fittest individual benefits from further training, even without an explicit form of adversarial defense. Considering the same threat model that was used to guide the evolutionary search, the clean and the adversarial accuracy against an FGSM adversary are boosted to around 93.47\% and 47.88\%, on average, after undergoing the extended training. Furthermore, the individual also shows some robustness against $L_2$-bounded perturbations, even if such samples were not seen during evolution: the accuracy on adversarial examples generated by FGM (the $L_2$ equivalent of FGSM) is around 58.35\%. Although the clean accuracy of the NERO-Net model is slightly lower when compared to the NSGA-Net model, the overall post-attack accuracy (i.e., the number of correctly classified samples taking the full test set into account instead of only $N_c$) of NERO-Net remains higher under both threat models (44.76\% vs. 37.97\% and 54.54\% vs. 52.55\% in $L_\infty$ and $L_2$, respectively).

\begin{table}
    \centering
    \caption{Clean and adversarial accuracy on CIFAR-10, with $\epsilon = 8/255$ in $L_\infty$ and $\epsilon = 0.5$ in $L_2$. Best results for standard and adversarial training are \underline{underlined} and in \textbf{bold}, respectively.}
    \label{tab:results}
\begin{minipage}{\columnwidth}
\footnotesize
    \centering
    \begin{tabular}{llcccc}
    \toprule
    \multirow{2}{*}{\textbf{Norm}}         & \multirow{2}{*}{\textbf{Attack}}                                    & \multicolumn{2}{c}{\textbf{Standard Training}}              & \multicolumn{2}{c}{\textbf{Adversarial Training}}     \\
    \cmidrule(lr){3-4} \cmidrule(lr){5-6}
                                           &                                                                     & \textbf{NSGA-Net}            & \textbf{NERO-Net}            & \textbf{NSGA-Net}    & \textbf{NERO-Net}    \\
    \midrule
    -                                      & None                                                                & \underline{95.55 $\pm$ 0.11} & 93.47 $\pm$ 0.15             & \textbf{84.89 $\pm$ 0.22} & 84.08 $\pm$ 0.40          \\
    \cmidrule{1-6}
    \multirow{3}{*}{$L_\infty$}       & FGSM                                                                & 39.73 $\pm$ 1.20             & \underline{47.88 $\pm$ 1.11} & \textbf{64.93 $\pm$ 0.59} & 60.09 $\pm$ 0.62          \\
                                           & PGD20                                                               & 00.00 $\pm$ 0.00             & \underline{00.13 $\pm$ 0.07} & \textbf{50.75 $\pm$ 0.74} & 44.69 $\pm$ 0.64          \\
                                           & AA                                                                  & -                            & -                            & \textbf{47.13 $\pm$ 0.86} & 40.40 $\pm$ 0.66          \\
    \cmidrule{1-6}
    \multirow{3}{*}{$L_2$}            & FGM                                                                 & 54.99 $\pm$ 0.83             & \underline{58.35 $\pm$ 0.72} & 76.25 $\pm$ 0.52          & \textbf{77.88 $\pm$ 0.39} \\
                                           & PGD20                                                               & 00.28 $\pm$ 0.04             & \underline{08.69 $\pm$ 0.58} & 64.45 $\pm$ 1.19          & \textbf{72.83 $\pm$ 0.49} \\
                                           & AA\footnote{Custom version with only APGD-CE and APGD-T.}           & -                            & -                            & 60.79 $\pm$ 1.31          & \textbf{70.98 $\pm$ 0.61} \\
    \bottomrule
    \end{tabular}
\end{minipage}
\vspace{-15pt}
\end{table}

Nevertheless, it would be misleading to say that our fittest individual is robust under these threat models, since the more powerful PGD adversary still brings its overall accuracy to near-zero values, with $L_\infty$ perturbations, and below 10\% (worse than a random classifier), in the case of $L_2$. The same happens with the NSGA-Net model, which reaches near-zero accuracies under both threat models.

Thus, we complement the models' inherent capability of resisting attacks by performing adversarial training.
We consider stronger adversaries in this scenario, namely the standard AA ensemble of attacks, to avoid overestimating robustness. Results are presented in Table~\ref{tab:results}.



As expected, adversarial training improves the robustness of our fittest individual in the $L_\infty$ scenario, although at the cost of clean accuracy, which drops to 84.08\% (previously 93.47\%), on average. If we focus on the attack that guided our evolutionary search, we see that adopting this defense allows for gains around 12 percentage points in comparison with the corresponding undefended model (60.09\% vs. 47.88\%).
Moreover, the adversarial accuracy after the complete AA ensemble now reaches, on average, 40.40\%.
We note that the last two attacks in AA (targeted FAB attack and Square attack) have a negligible contribution to the success of this adversary ($<0.1$ percentage points). Therefore, an ensemble with only the APGD attack on the CE loss (APGD-CE) followed by the targeted APGD attack on the DLR loss (APGD-T) would already give a good robustness estimate while saving computational resources.

Finally, if we also take the misclassifications of clean samples into account, the overall accuracy of the NERO-Net model is, on average, 33.96\%. We acknowledge that this result falls short of the one attained with the NSGA-Net model, whose overall accuracy after AA is, on average, around 40\%.

However, we hypothesize that there are two aspects that negatively impact the robustness of NERO-Net. On the one hand, the NERO-Net model (26.72 million parameters) is a considerably larger model than the NSGA-Net model (3.37 million parameters), and so, it makes sense that it requires more resources (both in terms of data and time) for its training to converge. On the other hand, we adopted the same adversarial training setting for the two models to promote a fair comparison, but it may be more suitable for smaller sized models. Thus, we argue that a personalized adversarial training setup could potentially allow the NERO-Net model to achieve better results. We conducted some preliminary experiments and, for the same model architecture and a standard training setting, we were able to evolve learning strategies that had a measurable impact on the adversarial accuracy of the models. Further validation is needed, but we find it safe to assume that the same would happen in an adversarial training setting, especially if we take into consideration the findings of other works in the literature (e.g., \citeauthor{peng_etal_2023}~\cite{peng_etal_2023}).

We also extend our analysis to $L_2$-robustness for the adversarially trained models. Initial assessments revealed similar trends regarding the contribution of each individual attack to the success of the AA adversary, and so, instead of running the standard ensemble, we ran a custom version with only two attacks (APGD-CE and APGD-T). Contrary to what happens in $L_\infty$, the NERO-Net model is more robust than the NSGA-Net one when it comes to perturbations bounded by the $L_2$-norm, with differences around 10 percentage points under AA. As such, guiding the evolutionary search towards $L_\infty$-robust models seems to provide the additional benefit of the evolved models being able to resist $L_2$-bounded attacks. We argue that this multi-threat robustness, and the superior results when a standard training method is adopted, compensates for the less impressive results of the adversarially trained NERO-Net model under $L_\infty$.

\section{Conclusion}
\label{sec:conclusion}
In this work, we propose NERO-Net, a neuroevolutionary approach based on Fast-DENSER that incorporates the adversarial robustness of the searched CNNs into the fitness function and evaluation of candidate solutions. Besides the modifications directly linked with assessing the quality of the individuals in a robustness-aware scenario, we also introduce changes at the inner level representation (enabling an evolutionary unit to encode blocks of layers instead of a single one) and the genotypic level responsible for encoding layer connectivity, this way allowing NERO-Net to explore a larger set of CNN architectures.

Our search space is more complex than those typically adopted in NAS-based approaches. However, we show that NERO-Net still succeeds in finding architectures that better resist adversarial attacks, especially in the absence of a specialized adversarial defense. Incorporating adversarial training into the final models further boosts their robustness, but the results do not quite match those of other models designed via NAS for all the threat models that were considered. We hypothesize that this shortcoming could be overcome by optimizing the adversarial training setting for our evolved models, which we leave for future work.

Another aspect that requires further analysis is the relationship between the robustness of a model trained with a standard method vs. adversarial training. It remains unclear whether the fittest model in a standard setting corresponds to the fittest model under a different optimization framework. Yet another challenge that we identified is the much more volatile response of a model, in terms of adversarial accuracy, to small changes at the genotypic level. Therefore, we find a more in-depth analysis of this relationship worth pursuing in future endeavors. Directly related to this issue is the study of the choices made during the evolutionary process and the analysis of emergent architectural patterns associated with enhanced robustness in the fittest individuals.

Moreover, the computational cost of NE continues to be a drawback, even more so when the optimization objectives go beyond the standard accuracy of the evolved models. The use of zero-cost proxies in fitness evaluation, specifically designed with adversarial robustness in mind \cite{feng_etal_2025}, is one alternative we intend to pursue in the future.

\begin{acks}
This work is funded by national funds through \grantsponsor{FCT}{FCT – Foundation for Science and Technology, I.P.}, within the scope of the research unit \grantnum{FCT}{UID/00326} - Centre for Informatics and Systems of the University of Coimbra.
It is also supported by the \grantsponsor{PRR}{Portuguese Recovery and Resilience Plan (PRR)}{} through project \grantnum{PRR}{C645008882-00000055}, Center for Responsible AI. The first author is also funded by the FCT under the individual grant \grantnum[https://doi.org/10.54499/UI/BD/151047/2021]{FCT}{UI/BD/151047/2021}.
\end{acks}

\bibliographystyle{ACM-Reference-Format}
\bibliography{references}

\end{document}